\documentclass[11pt]{article}
\usepackage{coling2018}
\usepackage{times}
\usepackage{url}
\usepackage{latexsym}
\usepackage{bbm}
\usepackage{amssymb}
\usepackage{graphicx}
\usepackage{subfigure}
\usepackage{amsmath}
\usepackage{color}
\usepackage{xcolor}
\usepackage{bm}
\usepackage[]{caption2} 
\usepackage{CJKutf8}

\slname{Chinese}

\title{Sequence-to-Sequence Learning for Task-oriented Dialogue with Dialogue State Representation}
\sltitle{\begin{CJK}{UTF8}{gbsn} 面向任务型对话中基于对话状态表示的序列到序列学习   \end{CJK}}

\author{Haoyang Wen, Yijia Liu, Wanxiang Che*, Libo Qin, Ting Liu\\
	Research Center for Social Computing and Information Retrieval, \\
	Harbin Institute of Technology, China \\
	{\tt \{hywen, yjliu, car, lbqin, tliu\}@ir.hit.edu.cn} \\\
}
\date{}

\begin{document}
\maketitle
\begin{abstract}
	Classic pipeline models for task-oriented dialogue system require explicit modeling the dialogue states and
hand-crafted action spaces to query a domain-specific knowledge base.
Conversely, sequence-to-sequence models learn to map dialogue history to the response in current turn
without explicit knowledge base querying. In this work, we
propose a novel framework that leverages the advantages of classic pipeline and
sequence-to-sequence models. Our framework models a dialogue state as a fixed-size distributed
representation and use this representation to query a knowledge base via an attention mechanism.
Experiment on Stanford Multi-turn Multi-domain
Task-oriented Dialogue Dataset shows that our framework significantly outperforms other
sequence-to-sequence based baseline models on both automatic and human evaluation.
\end{abstract}
\makesltitle
\begin{slabstract}
	\begin{CJK}{UTF8}{gbsn}
		面向任务型对话中，传统流水线模型要求对对话状态进行显式建模。这需要人工定义对领域相关的知识库进行检索的动作空间。相反地，序列到序列模型可以直接学习从对话历史到当前轮回复的一个映射，但其没有显式地进行知识库的检索。在本文中，我们提出了一个结合传统流水线与序列到序列二者优点的模型。我们的模型将对话历史建模为一组固定大小的分布式表示。基于这组表示，我们利用注意力机制对知识库进行检索。在斯坦福多轮多领域对话数据集上的实验证明，我们的模型在自动评价与人工评价上优于其他基于序列到序列的模型。
	\end{CJK}
\end{slabstract}
\blfootnote{
	\hspace{-0.65cm}  
	This work is licenced under a Creative Commons 
	Attribution 4.0 International Licence.
	Licence details:
	\url{http://creativecommons.org/licenses/by/4.0/}
}
\blfootnote{
	\hspace{-0.65cm}
	* Email corresponding.
}
\section{Introduction}
\begin{figure}[!tp]
	\centering
	\normalsize
	\label{fig1}
	\begin{tabular}{|l|l|l|l|l|}
		\hline\textbf{Address} & \textbf{Distance} & \textbf{POI type} & \textbf{POI} & \textbf{Traffic info} \\\hline
		638 Amherst St&3 miles&grocery store&Sigona Farmers Market&car collision nearby\\
		269 Alger Dr&1 miles&coffee or tea place&Cafe Venetia&car collision nearby\\
		5672 barringer street & 5 miles & certain address & 5672 barringer street & no traffic \\
		200 Alester Ave&2 miles&gas station&Valero&road block nearby\\
		899 Ames Ct&5 miles&hospital&Stanford Childrens Health&moderate traffic\\
		481 Amaranta Ave&1 miles&parking garage&Palo Alto Garage R&moderate traffic\\
		145 Amherst St&1 miles&coffee or tea place&Teavana&road block nearby\\
		409 Bollard St&5 miles&grocery store&Willows Market&no traffic\\\hline
	\end{tabular}
	\\
	\begin{tabular}{ll}
		\textbf{Driver:} & Address to the gas station.\\
		\textbf{Car:} & Valero is located at 200 Alester Ave.\\
		\textbf{Driver:} & OK , please give me directions via a route that avoids all heavy traffic.\\
		\textbf{Car:} & Since there is a road block nearby, I found another route for you and I sent it on your screen.\\
		\textbf{Driver:} & Awesome thank you.\\	
	\end{tabular}
	\caption{An example of a task-oriented dialogue that incorporates a knowledge base. The knowledge 
		base will be changed based on different dialogue environment setting. Agents need to generate 
		response based on current knowledge base.}
\end{figure}

	Task-oriented dialogue system attracts more and more attention with the success of commercial
systems such as Siri, Cortana and Echo. It helps users complete specific tasks with natural language. Figure \ref{fig1} shows a typical example of a task-oriented
dialogue, where an agent provides with location information for a user. The requirements for the
agents to accomplish users' demands usually involve querying the knowledge base (KB), like
acquiring address from location information KB in Figure \ref{fig1}.

Typical machine learning approaches model the problem as a partially observable Markov Decision 
Process (POMDP) \cite{williams-young:2007:CSL,young:2013:IEEE}, where a pipeline system is 
introduced. The pipeline system consists of four components: natural language understanding (NLU, Tur and De Mori, 2011)
\nocite{tur-demori:2011}, dialogue state tracking \cite{williams:2013:SIGDial,williams:2012:NAACL}, dialogue 
policy learning \cite{young:2010:CSL} and natural language generation \cite{wen:2015:ACL}. Taking the utterance in Figure \ref{fig1} for example,
NLU maps the utterance ``\texttt{Address to the gas station}'' into semantic slot ``\texttt{POI type}''.
Dialogue state tracker keeps the probability of ``\texttt{gas station}'' close to 1
against other values of slot ``\texttt{POI type}''.
Given a semantic frame as a dialogue state, which is the combination of distributions of these slots,
dialogue policy learning generates the next pre-defined system action, which usually involves querying the knowledge base.
The action is then converted to its natural language
expression using natural language generation. Both natural language understanding and dialogue state 
tracking require a large amount of domain specific annotation for training, which is expensive to 
obtain. Besides, the design of actions and the explicit forms of semantic frames require a lot 
of knowledge from human experts, which are domain-specific as well.

Neural generative models, typically Seq2Seq models, have achieved success on machine 
translation \cite{sutskever:2014:NIPS,bahdanau-cho-bengio:2014:arxiv,luong-pham-manning:2015:EMNLP}. This success spurs the 
interests to apply Seq2Seq models into dialogue systems. Seq2Seq models
map dialogue history directly into the response in current turn,
while requires a minimum amount of hand-crafting.
However, conventional Seq2Seq doesn't model the exterior data retrieval explicitly,
which makes it hard for Seq2Seq to generate information stored in KB like meeting time and address,
but this kind of retrieval is easy to achieve for classic pipeline.
To tackle with the problem, \newcite{eric-manning:2017:EACL} use an additional copy mechanism to retrieve entities that occurs in  both KB and dialogue history. 
\newcite{eric:2017:SIGDial} further introduced retrieval from key-value KB where the model uses key representations to retrieve the corresponding values.
However, not all KBs are presented in key-value forms. 
Besides, an important component of classic pipeline, dialogue state tracker,
is not properly modeled, making it difficult to precisely retrieve from KB.

In this paper, we propose a novel framework that takes the advantages from both classic pipeline 
models and Seq2Seq models. We introduce dialogue states into 
Seq2Seq learning, but in a implicit way. Distributions in classic state tracking are modeled as a group of representation vectors computed by an attention-based network \cite{britz-guan-luong:2017:EMNLP},
which can be considered as a dialogue state representation that aggregates information for each slot. And training this representation doesn't require annotation of dialogue state tracking.
Our model queries the KB entries in an 
attention-based method as well, so that the querying is differentiable, without domain-specific pre-defined action spaces. Meanwhile we compute the representation for KB using entry-level attention and aggregate the representation with dialogue state representation to form a memory matrix of dialogue history and KB information.
While decoding, we perform an attention over memory and an attention over input, incorporating copying mechanism \cite{gu:2016:ACL} that allows model to copy words from KBs to enhance the capability of retrieving accurate entities.

We evaluate the proposed framework on Stanford Multi-turn, Multi-domain Dialogue Dataset 
\cite{eric:2017:SIGDial}, to test the effectiveness of our framework and flexibility to apply to different 
domains. We compare our model with other Seq2Seq models and discovered that our 
model has outperformed other models on both automatic evaluation and human evaluation.

\section{Proposed Framework}
\begin{figure}[!tp]
\label{fig22}
 \includegraphics[width=\linewidth, trim={0 1cm 0 1.5cm}, clip]{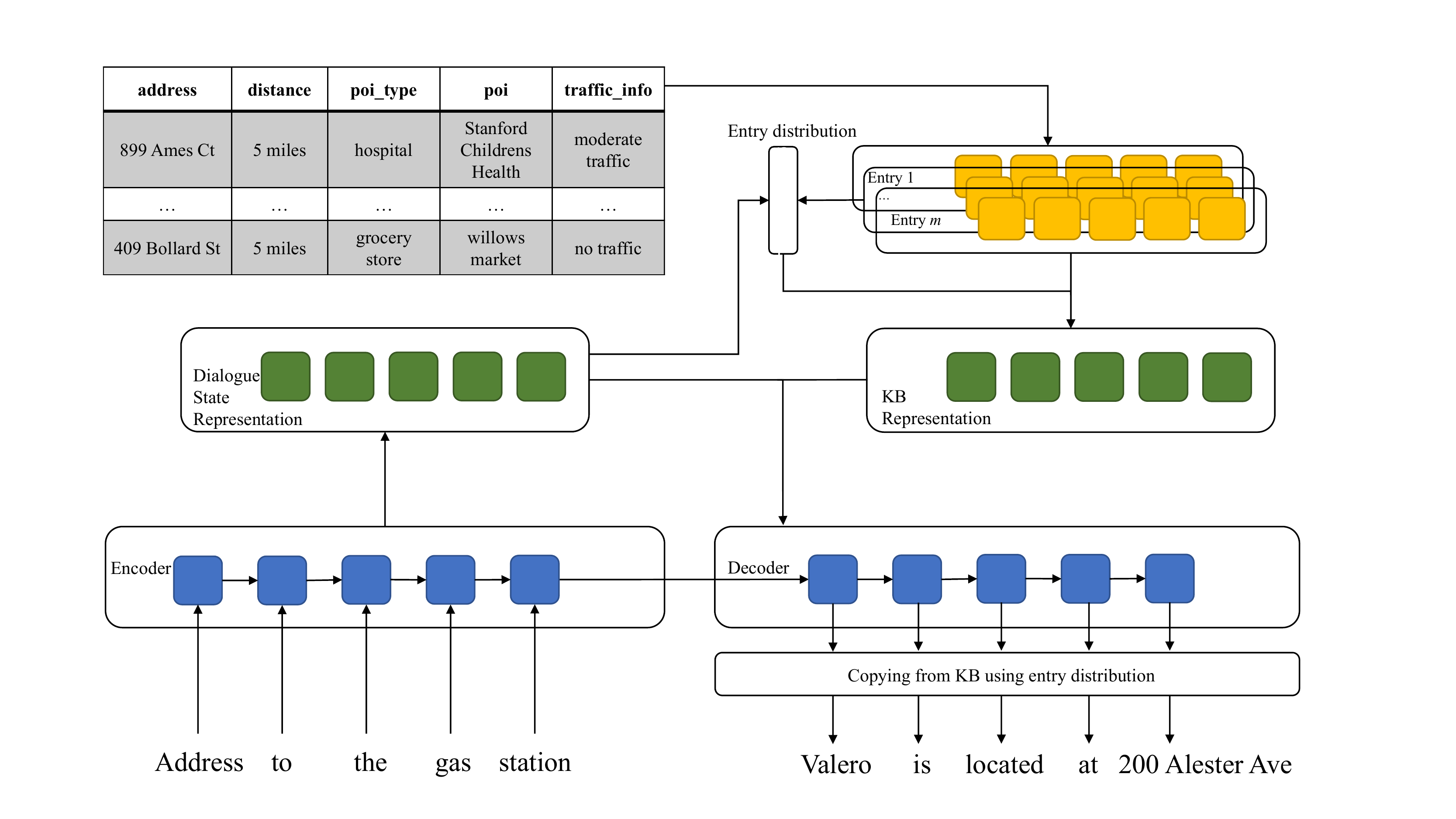}
 \caption{Proposed framework.}
\end{figure}

In this section, we describe a framework for task-oriented dialogue system. 
Our framework first encodes previous dialogue history, and computes dialogue state representation. 
Then our framework queries the table by attention and computes a matrix to represent information from previous history and KB. 
At last, the responses are generated using copying mechanism. 
The general architecture is demonstrated in Figure \ref{fig22}. 
\subsection{Encoder}
Given a dialogue consisting of utterances from a user and an agent, our encoder encodes the whole dialogue history. We represent the dialogue history
as a sequence of utterances. We encode the previous dialogue history as one single sequence 
consisting of each word in previous dialogue history and use $\left (x_1, x_2,\ldots, x_{n^\text{IN}} \right)$ to denote the whole dialogue history word by word, where $n^\text{IN}$ is the length of this sequence. Words are 
mapped to word embeddings and a long short-term memory network (LSTM) aggregates hidden representation 
over the sentence, denoted as $ \boldsymbol H^\text{ENC}=\left [\boldsymbol h^\text{ENC}_1, \ldots,
\boldsymbol h^\text{ENC}_{n^\text{IN}}\right]$ ($\boldsymbol H^\text{ENC} \in \mathbb{R}^{d\times
	n^\text{IN}}$, $d$ is the dimensionality of a hidden state).

\subsection{Dialogue State Representation}
The dialogue state tracking component of a dialogue system interprets the previous dialogue history to 
a belief state \cite{williams:2013:SIGDial}, which consists of a group of probability distributions over 
values for each slot. The dialogue state tracking is the core component of pipeline model. It helps with retrieving values from KB and generating accurate entities. To introduce dialogue state into Seq2Seq learning, in this framework, we model representation of belief state, motivated by 
\newcite{britz-guan-luong:2017:EMNLP}. We do not compute the explicit probability distribution for each 
slot. Instead, a group of distributed representations is computed. 
In this paper, we assume each turn of the dialogue is associated with $m$ slots and its dialouge state representation
is a matrix $\boldsymbol U^\text{IN} = \left[\boldsymbol u^\text{IN}_1, \ldots, \boldsymbol 
u^\text{IN}_m\right]\in \mathbb{R}^{d\times m}$ whose columns represent corresponding distribution.
We further assume that $m$ equals to the number of columns in our KB.
Each state representation $\boldsymbol{u}_k^{\text{IN}}$ is calculated 
by an attention over the whole representation of history:
\[
\boldsymbol u^\text{IN}_k\quad =\quad \sum_{t=1}^{n^\text{IN}}a^{\text{IN}}_{k}\left(t\right) \boldsymbol h^\text{ENC}_t,\\
\]
where we assign each hidden state $m$ different kinds of scores and perform a weighted average. The 
scores are computed by the following equations:
\begin{align*}
\textsc{ScoreIn}\left(\boldsymbol w^{\text A}_k, \boldsymbol h^\text{ENC}_t\right)\quad &=\quad \boldsymbol w^{\textit A^\textit T}_k \boldsymbol h^\text{ENC}_t,\\
a^{\text{IN}}_k\left(t\right)\quad  &=\quad \frac{\exp\left(\textsc{ScoreIn}\left(\boldsymbol w^{\text A}_k, \boldsymbol h^\text{ENC}_t\right)\right)}{\sum_{t'}\exp\left(\textsc{ScoreIn}\left(\boldsymbol w^{\text A}_k, \boldsymbol h^\text{ENC}_{t'}\right)\right)},
\end{align*}
$\boldsymbol W^{\textit A} = [\boldsymbol w^{\textit A}_1, \ldots, \boldsymbol w^{\textit A}_m]$ is a 
parameter matrix in $\mathbb{R}^{d\times m}$.

\subsection{Soft KB Attention}
\paragraph{Table Encoder.}
In our framework, we compile the process of querying  a KB entry into an attention network. The first step is to encode the tables.
Each KB cell is represented as the concatenation of the column name embedding $\phi^\textit{EMB}(f)$ and the cell value embedding $\phi^\textit{EMB}(c)$.
This representation is further fed into a $\tanh$ non-linearity and the final representation can be formalized as $\boldsymbol c = \tanh \left(\boldsymbol W^\text C \left[ \phi^\textit{EMB}\left(c\right), \phi^\textit{EMB}\left(f\right)\right]\right)$.
The representation of the KB entry $\boldsymbol{C}_k$ is denote as $\boldsymbol C_k = [\boldsymbol c_{k,1}, ..., \boldsymbol c_{k,m}]$, consisting all its cell representations.

\paragraph{Entry and KB Representation.}
\label{KB}
Conventionally, task-oriented dialogue systems interact with KB via carefully hand-crafted API calls, which are usually domain-specific and break the differentiability \cite{wen:2017:EACL}. To make it differentiable, our framework applies a soft-attention over the KB entries and the attention value can be interpreted as the possibility that an entry will be used for decoding. 
	We use the similarity between $\boldsymbol C_k$ and $\boldsymbol U^{\text{IN}}$ to represent attention score for the $k^{\rm th}$ entry. 
	The similarity is computed by the following equation:

\[
\text{sim}\left(\boldsymbol C_k, U^{\text{IN}}\right) \quad=\quad \sum_{t = 1}^{m}{\boldsymbol c_{k,t}\cdot \boldsymbol u^{\text{IN}}_{t}},
\]
where we sum the dot product between vectors in $\boldsymbol C_k$ and $\boldsymbol U^{\text{IN}}$ respectively. 

This similarity distribution is then converted into a probability distribution naturally using a softmax function. This probability distribution indicates the possibility to use entry $e_k$ in the further response generation given previous dialogue history $\boldsymbol x_{<i}$ :
\[
p(e = e_k\mid \boldsymbol x_{<i}) \quad=\quad \frac{\exp\left(\text{sim}\left(\boldsymbol C_k, U^{\text{IN}}\right)\right)}{\sum_{k'}\exp\left( \text{sim}\left(\boldsymbol C_{k'}, U^{\text{IN}}\right)\right)}.
\]

The information matrix from KB that is used for future generation can be computed as a weighted summation:
\[
\boldsymbol U^{\textit{KB}} \quad=\quad \sum_{t = 1}^{|T|}p(e = e_t\mid \boldsymbol x_{<i})  \boldsymbol C_t,
\]
where $\boldsymbol U^{\text{KB}}$ is denoted as the information matrix, $|T|$ is the number of entries in this KB.

Since the dimensionalities of all parameters are not related to the size of knowledge base. it allows changing the KB on-the-fly. Besides, there is no need to define specific operations, which is required for using API calls to extract information. Since we model the entries directly, it is appropriate to extract information from non-entity-centric knowledge bases as well.

Finally, we combine these two kinds of information by concatenating corresponding vectors and feeding them into a linear layer with an activation function. Formally, we denote $\boldsymbol U^{\text{CAT}}$ as the concatenation of $\boldsymbol U^{\text{IN}}$ and $\boldsymbol U^{\text{KB}}$. The two matrices are concatenated by concatenated corresponding vectors respectively. The process can be formulated as:
 $\boldsymbol U = \tanh \left(\boldsymbol W^{\textit{CAT}} \boldsymbol U^{\textit{CAT}}\right)$. The matrix $\boldsymbol{U}$ can be considered as a fix-sized memory representation over dialogue history and knowledge base.

\subsection{Decoder}
In this section, we will discuss the decoder that takes all previously calculated information to generate sentences. The vanilla Seq2Seq decoder generates a sequence of words recurrently based on the last hidden state of a encoder. We denote $\left(\boldsymbol h^{\text{DEC}}_1, \ldots, \boldsymbol h^{\text{DEC}}_{n^{\text{OUT}}}\right)$ as the hidden states of the decoder and $\left(y_1, \ldots, y_{n^{\text{OUT}}}\right)$ as the output sentence. We will consider two kinds of information, that are information of dialogue history and information from KB. The model aggregates information via attention over KB representation and history representation.
\paragraph{Input Attention.}
The conventional attention mechanism is introduced to extend the decoder, where each hidden state in the encoder is assigned a score based on the current hidden state $\boldsymbol h^{\text{OUT}}_t$ at time step $t$, and then the context vector is computed by the weight summation \cite{luong-pham-manning:2015:EMNLP}. This process can be described by the following equations:
\begin{align*}
\boldsymbol{c}^{\text{IN}}_t\quad &=\quad \sum_{i=1}^{n^{\text{IN}}}a^\text{OUT}_t\left(i\right)\boldsymbol h^{\text{ENC}}_i,\\
\textsc{ScoreOut}\left(\boldsymbol h^{\text{ENC}}_i, \boldsymbol h^{\text{DEC}}_t\right) \quad&=\quad \boldsymbol v^{\text{OUT}}\tanh\left(\boldsymbol W^{\text{OUT}}\left[\boldsymbol h^{\text{ENC}}_i, \boldsymbol h^{\text{DEC}}_t\right]\right).\\
a^\text{OUT}_t\left(i\right) \quad&=\quad \frac{\exp\left(\textsc{ScoreOut}\left(\boldsymbol h^{\textit{ENC}}_i, \boldsymbol h^{\text{DEC}}_t\right)\right)}{\sum_{i'}\exp\left(\textsc{ScoreOut}\left(\boldsymbol h^{\text{ENC}}_{i'}, \boldsymbol h^{\text{DEC}}_t\right)\right)}.\\
\end{align*}

\paragraph{Memory Attention.}
Besides the context vector through input attention, we also use another context vector from the attention over the fix-size memory matrix $\boldsymbol U$ and it's computed as:
\begin{align*}
\boldsymbol{c}^{\text{MEM}}_t \quad&=\quad \sum_{i=1}^{m}a^\text{MEM}_t\left(i\right)\boldsymbol u_i,\\
\textsc{ScoreMem}\left(\boldsymbol u_i, \boldsymbol h^{\text{DEC}}_t\right)\quad&=\quad \boldsymbol v^{\text{MEM}}\tanh\left(\boldsymbol W^{\text{MEM}}\left[\boldsymbol u_i, \boldsymbol h^{\text{DEC}}_t\right]\right),\\
a^{\text{MEM}}_t\left(i\right) \quad&=\quad \frac{\exp\left(\textsc{ScoreMem}\left(\boldsymbol u_i, \boldsymbol h^{\text{DEC}}_t\right)\right)}{\sum_{i'}\exp\left(\textsc{ScoreMem}\left(\boldsymbol u_{i'}, \boldsymbol h^{\text{DEC}}_t\right)\right)}.\\
\end{align*}
In practice, we use an additional context vector from encoder to calculate the dialogue state representation, which is different from the one that is used for the start of decoding.

The two kinds of context vectors and the current hidden state are used for decoding. 
We introduce a variant of copying mechanism \cite{gu:2016:ACL} in order to retrieve entities from KB directly. We first compute a probability distribution over output vocabulary $\mathcal{V}$ and slot types $\mathcal{V}^{\text{SLOT}}$, given previous dialogue history $\boldsymbol x_{<i}$ and previous generated words $\boldsymbol y_{<t}$, which can be described as:
$p\left(\widetilde{y_t}\mid \boldsymbol x_{<i}, \boldsymbol y_{<t}\right)
= \mathrm{softmax}\left(W^{\text O}\left[\boldsymbol h^{\text{DEC}}, \boldsymbol c^{\text{IN}}, \boldsymbol c^{\text{MEM}}\right]\right).$
The probability of generating a slot type represents the sum of probability of generating a slot value for this slot from KB. Since we have calculated the probability of using an entry in section \ref{KB}, the probability of generating a word can be described as:
\begin{align*}
p\left(y_t = y\mid\widetilde{y_t}, \boldsymbol x_{<i}, \boldsymbol y_{<t}\right) \quad=\quad &p\left(\widetilde{y_t} = y\mid \boldsymbol x_{<i}, \boldsymbol y_{<t}\right) \\&+ \sum_{y^{\text{S}} \in \mathcal V^{\text{SLOT}}} p\left(\widetilde{y_t} = y^{\text S}\mid \boldsymbol x_{<i}, \boldsymbol y_{<t}\right) \sum_{k = 1}^{|T|}\mathbbm{1}\left\{e_k\left(y^{\text S}\right) = y\right\}p(e = e_k\mid \boldsymbol x_{<i}),
\end{align*}
where $e_k\left(y^\text{S}\right)$ is the cell that is in slot (i.e. column) $y^\text{S}$.
Note that the summation of probability over $\mathcal{V}$ is exactly 1 after the model copies entities from KB.

\subsection{Training}
Conventionally, we use negative log likelihood (NLL) for training to train a Seq2Seq model. 
Since there is no supervision for soft KB attention, and it is easy to get over-fitting when we only use NLL,  we apply policy gradient to improve the performance of soft KB attention as well. We consider the KB and fix-size memory representation from input as the environment in a reinforcement learning setting. There is only one action, which is defined as choosing a single entry from KB to help with generating response. Heuristically, the more entities in an entry appear in dialogue context, the higher possibility that this entry is used for generating response. Therefore, we consider the number of entities from an entry $e$ that appear in previous dialogue history or current gold response as reward $R\left(e\right)$, and apply REINFORCE with baseline \cite{williams:1992:RL}:
$J_{\textit{RL}} = -\mathbb{E}_{p(e\mid \boldsymbol x_{<i})}\left[R\left(e\right) - b\right]$,
where $b$ is a hyperparameter denoting the baseline reward. The joint loss is the combination of the NLL and loss from reinforcement learning, which is:
\begin{align*}
J \quad=\quad J_{\textit{NLL}} + \lambda J_{\textit{RL}}
\end{align*}
where $\lambda$ is a hyperpramater balancing the two factors. In practice, we first use $J_{\textit{RL}}$ only to train the soft KB attention and its encoder, without training for response generation, which will accelerate the convergence of Seq2Seq learning. We apply data augmentation to the original dataset as well, where we add delexicalised form responses into training data in order to force our model to generate slot types first and then retrieve entity from KB using copying mechanism. The delexicalised responses are generated by simple matching and replacing.

\section{Experiment}
In this section, we first introduce the details of experiment setting. Then we provide results and analyses of automatic evaluation and human evaluation in order to compare with other baseline models. Besides, we present ablation test to evaluate and analyze the function of different components in our framework. Finally, we provide visualization of dialogue state representation and case study.
\begin{table}
	\label{table1}
	\begin{tabular}{l|p{6.5cm}|p{6.5cm}}
		&\multicolumn{1}{c|}{\textbf{Weather}}&\multicolumn{1}{c}{\textbf{Navigation}}\\\hline
		\textbf{Slot Types}& location, date, highest temperature, lowest temperature and weather attribute&POI name, traffic info, POI type, address, distance\\
	\end{tabular}
	\caption{Slot types for different domains.}
\end{table}
\subsection{Experiment Setting}
We choose two KB-rich domains from Stanford Multi-turn Multi-domain Task-oriented Dialogue Dataset  \newcite{eric:2017:SIGDial}, which are weather information retrieval and point-of-interest (POI) navigation (navigation). We first change the form of KB in weather information retrieval domain (weather). \newcite{eric:2017:SIGDial} integrates the highest temperature, lowest temperature and weather information into a single weekday column due to their incapability of utilizing non-entity-centric KB. In this paper, we separate these information into three different column, and the slots of these two domains are provided in Table \ref{table1}.

Our framework is trained separately in these two domains, using the same train/validation/test split sets as \newcite{eric:2017:SIGDial}. We do not map the entities in dialogue into its canonical form as what \newcite{eric:2017:SIGDial} have done, since our framework extract entities directly from KB. And we evaluate our framework on exact entities as well.

Our framework is trained using the Adam optimizer \cite{kingma-ba:2014:ICLR}. The learning rate is $10^{-3}$, $\lambda$ for balancing two loss functions is $10^{-1}$. We applied dropout \cite{srivastava:2014:JMLR} to the input and the output of LSTM, with a dropout rate at $0.75$. We add the weight decay on the model. The coefficient of weight decay is $5\times 10^{-6}$. The embedding size and all hidden size are 200. The number of epochs for training soft KB attention is 30 for both navigation and weather. Baseline $b$ is 1.5 for both navigation and weather.
\subsubsection{Baseline Models}
We compare our model with two additional baselines beyond the Seq2Seq with attention, which includes:
\begin{itemize}
\item \textbf{Copy-augmented Sequence-to-Sequence Network.} This model is adapted from \cite{eric-manning:2017:EACL}. It utilizes entities that appear in both previous dialogue history and KB. A hard copy mechanism for these entities is applied in this model.
\item \textbf{Key-value Retrieval Network.} This model is adapted from \newcite{eric:2017:SIGDial}. It utilizes key-value forms to represent KBs. Key representations are used for an attention-based value retrieval. In weather information retrieval domain, although we have changed the KB into an non-entity-centric form, we still designate ``\texttt{location}'' slot as subject slot and we allow a key representation to retrieval multiple values. We convert inputs into canonical forms, while the outputs remain the same in order to compare with our model.
\end{itemize}

\subsection{Automatic Evaluation}
In this section, we provide two different automatic evaluations to compare with other baseline models. The results and analyses are provide in the following sections.

\subsubsection{Evaluation Metrics}

Entity F1 and BLEU score are used to evaluate our model.
	The entity F1 scores in both micro-average and macro-average manners are used to measure the difference between entities in the system and gold responses. Besides, we use BLEU to evaluate the quality of responses. \newcite{sharma:2017:arxiv} showed that BLEU has a comparatively strong correlation with human evaluation on task-oriented dialogue dataset. The BLEU score is computed from results with highest micro F1. To evaluate macro F1, we delete instances that neither gold nor generated response contains an entity.

\subsubsection{Results and Analyses}
Experiment results are illustrated in Table \ref{table2}. 
The results show that our model outperforms other models in most of automatic evaluation metrics. 
In the navigation domain, compared to KV Net, we achieve 5.0 improvement on BLEU score, 37.1 improvement on Macro F1 and 27.4 improvement on Micro F1. 
Compared to Copy Net, we achieve 5.0 improvement on BLEU score, 41.2 improvement Macro F1 and 27.4 improvement on Micro F1. 
The results in navigation show our model's capability to generate more natural and accurate response than the Seq2Seq baseline models. 
In the weather domain, our model generates more accurate responses than our baseline models as well. 
The BLEU score is a little bit lower than Copy Net and Seq2Seq with attention. This is because the forms of responses are relatively limited in weather domain.
Besides, the entities in inputs are highly probable to be mentioned in responses, such as ``\texttt{location}''. These two reasons indicate that the simpler models can capture this pattern more smoothly.
The results that Seq2Seq with Attention performs better than Copy Net and KV Net also confirm this.

We also find that the KV Net's results are lower than that reported by \newcite{eric:2017:SIGDial}.
	We address this to the differences in the preprocessing, model training and evaluation metrics. 
	In spite of the difference of evaluation metrics that we evaluate on exact entities rather than their canonical forms, the Micro F1 score of our model still outperforms what \newcite{eric:2017:SIGDial} reported, which is 41.3 in navigation domain and which is evaluated on canonical forms.
	Our changes of the weather domain into non-entity-centric also influence its performance.
	This differences in results also indicate the robustness of our model when facing non-entity-centric KBs.
\begin{table}
	\centering
	\begin{tabular}{l|ccc|ccc}
		&\multicolumn{3}{c|}{\textbf{Navigation}}&\multicolumn{3}{c}{\textbf{Weather}}\\
		\textbf{Model}&\textbf{BLEU}&\textbf{Macro F1}&\textbf{Micro F1}&\textbf{BLEU}&\textbf{Macro F1}&\textbf{Micro F1}\\\hline
		Seq2Seq with Attention&8.3&15.6&17.5&\textbf{19.6}&56.0&53.5\\
		Copy Net &8.7&20.8&23.7&17.5&52.4&53.1\\
		KV Net &8.7&24.9&29.5&12.4&37.7&39.4\\
		our model&\textbf{13.7}&\textbf{62.0}&\textbf{56.9}&14.9&\textbf{58.5}&\textbf{56.3}\\
	\end{tabular}
	\caption{Automatic evaluation on test data. Best results are shown in bold. Generally, our framework outperforms other models in most automatic evaluation metrics. }
	\label{table2}
\end{table}
\begin{table}[!tp]
	\centering
	\begin{tabular}{l|ccc}
		\textbf{Model}&\textbf{BLEU}&\textbf{Macro F1}&\textbf{Micro F1}\\\hline
		our model&\textbf{13.7}&\textbf{62.0}&\textbf{56.9}\\
		-copying&9.6&35.2&41.3\\
		-RL&9.3&38.2&46.0\\
	\end{tabular}
	\caption{Ablation experiment on navigation domain. 
		-copy refers to a framework without copying. -RL refers to a framework without RL loss.}
	\label{table3}
\end{table}
\subsubsection{Ablation}

In this section, we perform several ablation experiments to evaluate different components in our framework on the navigation domain. Results are shown in Table \ref{table3}. The results demonstrate effectiveness of components of our model to the final performance. 

Copying mechanism enables our framework to retrieve entities directly from KBs. Without copying mechanism, such retrieval is infeasible and our framework cannot produce values in KBs. The results show that it introduces more variability to the generation process if we do not use copying mechanism.

The reinforcement learning loss helps our framework to use correct KB entries so that improve the performance of generation. Without this reinforcement learning loss, the item selection process is only supervised by log likelihood loss. We address the drop in performance to that our model overfits to the training data.

\begin{table}[!tp]
	\centering
		\begin{tabular}{l|ccc}
	\textbf{Model}&\textbf{Correct}&\textbf{Fluent}&\textbf{Humanlike}\\\hline
	Copy Net &3.52&4.47&4.17\\
	KV Net&3.61&4.50&4.20\\
	our model&\textbf{4.21}&\textbf{4.65}&\textbf{4.38}\\\hline
	agreement&41.0&55.0&43.0
\end{tabular}
\caption{Human evaluation of responses based on random selected previous dialogue history in test dataset. The agreement scores indicate the percentage of responses to which all three human experts give exactly the same scores.}
\label{table4}
\end{table}

\subsection{Human Evaluation}
In this section, we provide human evaluation on our framework and other baseline models. We randomly generated 200 responses. These response are based on distinct dialogue history in navigation test data. We hire three different human experts to evaluate the quality of responses. Three dimensions are involved, which are correctness, fluency, and humanlikeness. Three human experts judged each dimension on a scale from 1 to 5. And each judgment indicates a relative score compared to standard response from test data. The results are illustrated in Table \ref{table4}. The results show that our framework outperforms other baseline models on all metrics. The most significant improvement is from correctness, indicating that our model generates more accurate information that the users want to know.

\begin{table}

\end{table}
\subsection{Visualization and Case Study}
We provide a visualization example to demonstrate the effectiveness of dialogue state representation. 
The visualization illustrates attention scores over the sentence for each slot. The blue-level of a cell indicates the attention score it represents.
From this visualization, we can discover that our dialogue state representation matches slots with correct entities in sentence. 
For example, ``\texttt{pizza\_restaurant}'' matches ``\texttt{poi}'' and 
``\texttt{poi\_type}'' correctly, ``\texttt{4\_miles}'' matches ``\texttt{address}'' correctly. The visualization indicates the capability of our framework to track accurate information and integrate them into a fix-size matrix representation.

Finally, we provide a case study that consists of two conversations which are generated between our framework and a human and between Seq2Seq with Attention with human respectively. In this case, we find that our framework is able to generate correct information such as address and point-of-interest. Conversely, Seq2Seq with Attention is generated in a more random way. The comparison between these two dialogues illustrates the capability of our model to retrieve accurate entities and while in the same time generate natural response.

\begin{figure}[!tp]
\centering
\includegraphics[scale=0.8, trim={0cm 1.3cm 0cm 0cm}, clip]{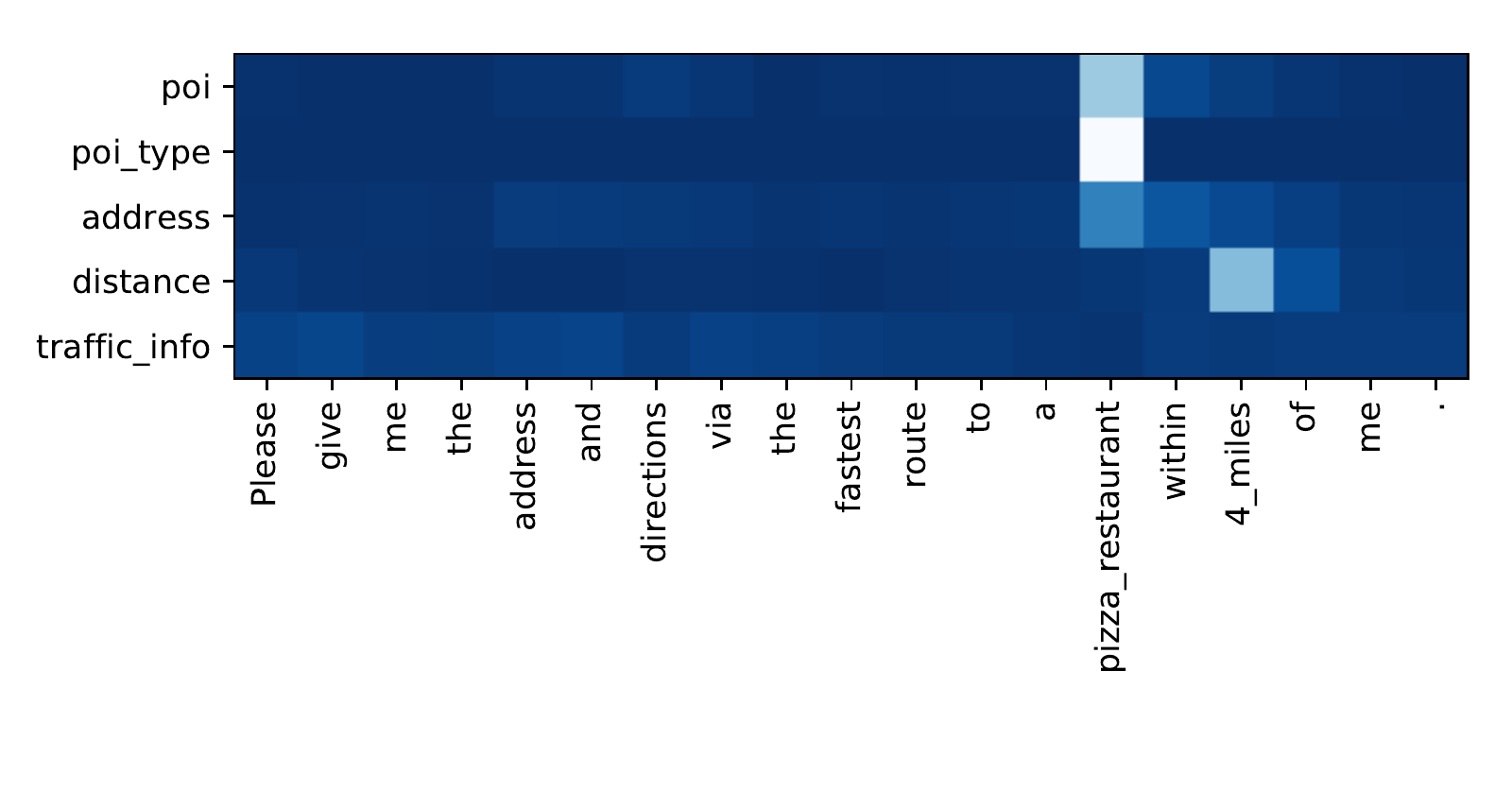}
\caption{Visualization of dialogue state representation.}
\end{figure}
\begin{figure}[!t]
	\centering
	\normalsize
	\label{fig4}
	\begin{tabular}{|l|l|l|l|l|}
		\hline\textbf{Address} & \textbf{Distance} & \textbf{POI type} & \textbf{POI} & \textbf{Traffic info} \\\hline
		783 Arcadia Pl&5 miles&gas station&Chevron&moderate traffic\\
		383 University Ave&5 miles&shopping center&Town and Country&no traffic\\
		864 Almanor Ln & 5 miles &friends house& jacks house & no traffic \\
		5671 barringer street&6 miles&home&home&heavy traffic\\
		657 Ames Ave&4 miles&rest stop&The Clement Hotel&no traffic\\
		638 Amherst St&1 miles&grocery store&Sigona Farmers Market&heavy traffic\\
		830 Almanor Ln&6 miles&chinese restaurant&tai pan&no traffic\\\hline
	\end{tabular}
	\\
	
	\begin{minipage}{0.49 \textwidth}
		\raggedright
		
		\begin{tabular}{lp{5.7cm}}
			\multicolumn{2}{c}{\textbf{our model}}\\
			\textbf{Driver:} & what gas stations are here ?\\
			\textbf{Car:} & there is a chevron 5 miles away .\\
			\textbf{Driver:} & that 's good ! please show me the address and pick the quickest route to get there !\\
			\textbf{Car:} & the address is 783 arcadia pl , i sent on your screen the best route to reach there .\\
			\textbf{Driver:} & thank you !\\
			\textbf{Car:} & you 're welcome ! \\
		\end{tabular}
	\end{minipage}
	\begin{minipage}{0.49\textwidth}
		\raggedleft
		\begin{tabular}{lp{5.7cm}}
			\multicolumn{2}{c}{\textbf{Seq2Seq with Attention}}\\
			\textbf{Driver:} & what gas stations are here ?\\
			\textbf{Car:} & there is a chevron 4 miles away .\\
			\textbf{Driver:} & that 's good ! please show me the address and pick the quickest route to get there !\\
			\textbf{Car:} & the address to safeway is 452 arcadia pl .\\
			\textbf{Driver:} & thank you !\\
			\textbf{Car:} & you 're welcome ! \\
			&\\
		\end{tabular}
	\end{minipage}
	\caption{Case study. The dialogue in the left is generated from our framework, and that in the right is generated from Seq2Seq with Attention baseline.}
\end{figure}
\section{Related Work}
The recent successes on neural networks spread across many natural language processing tasks, which also stimulate research on task-oriented dialogue system. 
The powerful distributed representation ability of neural networks makes task-oriented dialogue system end-to-end possible. 
Recently, \newcite{wen:2017:EACL} built a system that connects classic pipeline modules by a policy network training with a user simulator. \newcite{wen:2017:ICML} further introduced latent intention into end-to-end learning.
However, their modules like belief tracker still needs to be trained separately before end-to-end training. 
In contrast to their work, our framework trained the state tracker jointly with the end-to-end dialogue training. \newcite{liu:2017:interspeech} built a turn-level LSTM to model the dialogue state and generate probability distribution for each slot.
 \newcite{bordes-weston:2017:ICLR} built a system by applying memory network to store the previous dialogue history. But the responses are retrieved from templates, which is significantly different from our neural generative responses. Another type of work tried to build an end-to-end system as a task completion dialogue system \cite{li:2016:arxiv,li:2017:IJCNLP,peng:2017:EMNLP}. These modeled are trained through an end-to-end fashion via reinforcement learning algorithm using the reward from user simulators. However, their dialogue responses are not generated from the dialogue history directly but from a set of pre-defined action and explicit semantic frames.
 
Our soft KB attention can be considered as a process to retrieve entries, which has been explored in many QA and dialogue work.
One line of this research includes creating well-defined API calls to query the KB \cite{williams-asadi-zweig:2017:ACL,wen:2017:ICML}.
And another line of research tried to directly retrieve entities from knowledge base.
 \newcite{yin:2016b:IJCAI} has built a system to encode all table cells and assign a score vector to each row. Our framework resembles the second line of research, but can generate multiple entities to form natural language responses. 
\newcite{he:2017:ACL} has built two symmetric dialogue agents with private knowledge, and has applied knowledge graph reasoning into Seq2Seq learning, which is distantly related with our framework.
 In the sense of the KB forms, \newcite{yin:2016a:IJCAI} retrieved entities based on (\textit{subject, relation, object}) triples. 
 While \newcite{dhingra:2017:ACL} applied a soft-KB lookup on an entity-centric knowledge base to compute the probability of that the user knows the values of slots, and has tried to model the posterior distributions over all slots. 
 However, our framework doesn't require entity-centric knowledge base.

\section{Conclusion}
In this paper, we proposed a framework that leverages dialogue state representation, which is tracked by an attention-based methods. Our framework performed an entry-level soft lookup over the knowledge base, and applied copying mechanism to retrieve entities from knowledge base while decoding. This framework was trained in an end-to-end fashion with only the dialogue history, and get rid of other annotation. Experiments showed that our model outperformed other Seq2Seq models on both automatic and human evaluation. The visualization and case study demonstrated the effectiveness of dialogue state representation and entity retrieval.

\section{Acknowledgments}
We thank the anonymous reviewers for their helpful comments and suggestions.
This work was supported by the National Key Basic Research
Program of China via grant 2014CB340503 and the
National Natural Science Foundation of China (NSFC) via
grant 61632011 and 61772153.

\end{document}